\pdfoutput=1

\documentclass[11pt]{article}

\usepackage[]{ACL2023} 

\usepackage{times}
\usepackage{latexsym}
\usepackage[T1]{fontenc}
\usepackage[utf8]{inputenc}
\usepackage{microtype}
\usepackage{inconsolata}
\usepackage{tabularx}
\usepackage[frozencache,cachedir=.]{minted}

\title{Assessing the efficacy of large language models in generating accurate teacher responses}

\author{
    Yann Hicke, Abhishek Masand, Wentao Guo, Tushaar Gangavarapu\\
    Cornell University
}

\begin{document}
\maketitle

\begin{abstract}
\cite{tack_bea_2023} organized the shared task hosted by the 18th Workshop on Innovative Use of NLP for Building Educational Applications on generation of teacher language in educational dialogues. Following the structure of the shared task, in this study, we attempt to assess the generative abilities of large language models in providing informative and helpful insights to students, thereby simulating the role of a knowledgeable teacher. To this end, we present an extensive evaluation of several benchmarking generative models, including GPT-4 (few-shot, in-context learning), fine-tuned GPT-2, and fine-tuned DialoGPT. Additionally, to optimize for pedagogical quality, we fine-tuned the Flan-T5 model using reinforcement learning. Our experimental findings on the Teacher-Student Chatroom Corpus subset indicate the efficacy of GPT-4 over other fine-tuned models, measured using BERTScore and DialogRPT. 

We hypothesize that several dataset characteristics, including sampling, representativeness, and dialog completeness, pose significant challenges to fine-tuning, thus contributing to the poor generalizability of the fine-tuned models. Finally, we note the need for these generative models to be evaluated with a metric that relies not only on dialog coherence and matched language modeling distribution but also on the model's ability to showcase pedagogical skills.
\end{abstract}

\section{Introduction}
\label{sec:1}

The advent of powerful open-source generative language models such as GPT-2 \cite{radford2019language}, T5 \cite{raffel2020exploring}, OPT \cite{zhang2022opt}, BLOOM \cite{scao2022bloom}, Flan-T5 \cite{chung2022scaling} or LLAMA \cite{touvron2023llama} has led to significant developments in conversational agents, opening avenues for various applications in education \cite{wollny2021we}. Such AI-driven educational dialogues offer the potential for skill improvement and personalized learning experiences, with intelligent tutoring systems increasingly gaining traction \cite{bibauw2022dialogue}. However, deploying AI-based teachers in the educational ecosystem demands the careful modeling and evaluation of these agents to ensure their capability to address critical pedagogical concerns.
\newline
 \cite{tack2022ai} created the AI teacher test challenge which follows the recommendations from \cite{bommasani2021opportunities} (pp. 67-72) stating that, if we want to put generative models into practice as AI teachers, it is imperative to determine whether they can (a) speak to students like a teacher, (b) understand students, and (c) help students improve their understanding.
 \newline
 Taking inspiration from the AI teacher test challenge which asks whether state-of-the-art generative models are good AI teachers, capable of replying to a student in an educational dialogue this paper seeks to investigate the applicability of reinforcement learning (RL) techniques in the generation of AI teacher responses within educational dialogues. The AI teacher test challenge emphasizes the need for a systematic evaluation of generative models to ensure that they can effectively communicate with students, comprehend their needs, and facilitate their academic improvement. Can we guide the language generator with RL to help it focus on these pedagogical requirements? 
\newline
\cite{tack_bea_2023} organized the shared task hosted by the 18th Workshop on Innovative Use of NLP for Building Educational Applications on generation of teacher language in educational dialogues. Following the structure of the shared task, in this study, we aim to evaluate the potential of combining state-of-the-art generative language models with reinforcement learning algorithms to generate AI teacher responses in the context of real-world educational dialogues sourced from the Teacher Student Chatroom Corpus \cite{caines2020teacher, caines2022teacher}. The natural baselines for the task at hand are SOTA closed-source models such as GPT-4, and fine-tuned open-source pre-trained models such as GPT-2 \cite{radford2019language}. We will evaluate these natural baselines before evaluating fine-tuned pre-trained models using RL techniques, that optimize for pedagogical quality.
\newline
By exploring the role of reinforcement learning in guiding the generation of AI teacher responses, we aim to advance the discourse on the utilization of conversational agents in educational settings and contribute innovative ideas to the ongoing shared task on the generation of teacher language in educational dialogues at the 18th Workshop on Innovative Use of NLP for Building Educational Applications.
\newline
The rest of this paper is structured as follows. Section 2 offers a comprehensive review of relevant literature in the areas of AI-driven educational dialogues and reinforcement learning-based language generation. Section 3 discusses the analysis and processing of the dataset prior to conducting any language modeling tasks. In Section 4, the proposed model and its methodology for generating AI teacher responses in educational interactions are introduced. Section 5 evaluates the effects of our approach on the quality and relevance of the generated AI teacher responses and highlights key observations. Finally, Section 6 concludes the paper and explores potential directions for future research.

\section{Related Work}
\label{sec:2}
A variety of related literature exists in the realm of conversational teaching between a student and a teacher. In this section, we review several notable works addressing aspects of teacher-student dialogues, foundation models, and conversational datasets, which have contributed to the progress and understanding of generative models in educational contexts.
\newline
\newline
\textbf{Teacher-Student Dialogues}
\newline
One prominent resource in educational dialogues is the National Council of Teachers of English (NCTE) dataset \cite{Kane2022}. It includes numerous examples of teacher-student interactions, which can serve as a valuable resource for the training and evaluation of generative models in an educational context.
\newline
The SimTeacher dataset \cite{cohen2020teacher} is an assemblage of information obtained through a "mixed-reality" simulation platform. This unique environment aids beginner educators in honing essential skills for classroom settings by employing student avatars managed by human actors. All aspiring teachers from a prominent public university participate in several brief simulation sessions throughout their educational preparation program, focusing on improving their ability to encourage more profound textual understanding among students. The original researchers annotated a variable called "quality of feedback" within the transcript, determining how effectively teachers proactively assist students.
\newline
In \cite{chen2019predictors}, we can find a dataset collected from an education technology company that provides on-demand text-based tutoring for math and science. With a mobile application, a student can take a picture of a problem or write it down and is then connected to a professional tutor who guides the student to solve the problem. The dataset represents, after some selection, 108 tutors and 1821 students. Each session is associated with two outcome measures: (1) student satisfaction scores (1-5 scale) and (2) a rating by the tutor manager based on an evaluation rubric (0-1 scale).
\newline
\newline
\textbf{Foundation Models}
\newline
\cite{bommasani2021opportunities} provided a comprehensive analysis of the opportunities and risks of foundation models, including insights into their use in educational applications. They identified potential benefits, such as personalized learning and accessibility, while also highlighting the major risks, such as unfair biases and the generation of harmful content. This work establishes the need for carefully crafted benchmarks and evaluations to assess the potential of generative models in education.
\newline
The AI Teacher Test \cite{tack2022ai} builds on this idea by examining the performance of generative models such as GPT3 \cite{brown2020language} and Blender \cite{roller2020recipes} in generating appropriate and informative responses in a teacher-student dialogue.
\newline
Kasneci et al. \cite{kasneci2023chatgpt} conducted an investigation to understand the effectiveness of ChatGPT \cite{team2022chatgpt} as a tool for educational support. They analyzed the model's performance in a student-tutoring context, examining its ability to provide accurate, relevant, and engaging responses for learners. By identifying the strengths and weaknesses of ChatGPT in this specific setting, they contributed to a better understanding of how generative models can be successfully deployed in educational applications.
Our work builds on these foundations by evaluating the potential of combining reinforcement learning with generative models to enhance the performance of AI teacher agents in educational dialogues.
\newline
\newline
\textbf{Conversational Uptake}
\newline
\cite{collins1982discourse} introduced the concept of uptake as a way to comprehend the effectiveness of conversational responses in a teacher-student dialogue. It laid the groundwork for the evaluation of generative models in dialogues by taking into account the relevance and appropriateness of model-generated responses.
\newline
Demszky et al. \cite{demszky2021measuring} further explored the concept of Conversational Uptake by proposing metrics to assess the success of responses in maintaining and advancing a conversation. By applying these metrics to AI-generated responses, their work contributes to the evaluation of models in realistic conversation settings, including teacher-student dialogues. Our work attempts to guide the language generation process with similar goals in mind. We hope to find proxies of pedagogical quality through NLP metrics such as BERTScore combined with DialogRPT.

\indent We continue by reviewing the literature utilizing reinforcement learning as a guide for language generation.
\newline
\newline
\textbf{Reinforcement Learning for language generation}
\newline
Policy gradient-based algorithms and their variants have been widely used in text generation to optimize sequence-level metrics \cite{ranzato2015sequence, shen2015minimum, norouzi2016reward, pasunuru2018multi}. Off-policy Reinforcement Learning (RL) is also commonly used in dialogue applications where online interaction with users is expensive \cite{serban2017deep, jaques2019way}. The main difference in our work is that we take advantage of demonstrations and design generic reward functions for generation tasks.
We extend this concept to educational contexts by employing reinforcement learning to guide the generation of AI teacher responses in educational dialogues. We focus on optimizing the responses of fine-tuned generative models based on a reward system designed to enhance the pedagogical quality of the generated responses. Recently, Ramamurthy et al. \cite{ramamurthy2022reinforcement} explored the efficacy of using RL to optimize language models in several natural language processing tasks, including text classification, sentiment analysis, and language generation. They developed a library, RL4LMs, which provides a generic framework for deploying RL-based language models for various tasks. We build on top of the RL4LMs framework by adding a new task to its existing array of tasks which we hope can be added as a standard for any future RLHF benchmark.

\section{Data}
\label{sec:3}

The shared task for BEA 2023 is based on the Teacher-Student Chatroom Corpus (TSCC) \cite{caines2020teacher}. This corpus comprises data collected from 102 chatrooms where English as a Second Language (ESL) teachers interact with students to work on language exercises and assess the students' English language proficiency.

\subsection{Data Extraction and Format}

From each dialogue in the TSCC, several shorter passages were extracted. Each passage is at most 100 tokens long, consisting of several sequential teacher-student turns (i.e., the preceding dialogue context) and ending with a teacher utterance (i.e., the reference response). These short passages are the data samples used in this shared task.

The data samples are formatted using a JSON structure inspired by the ConvoKit \cite{chang2020convokit}. Each training sample is represented as a JSON object with three fields:

\begin{itemize}
    \item \textbf{id}: a unique identifier for the sample.
    \item \textbf{utterances}: a list of utterances corresponding to the preceding dialogue context. Each utterance is a JSON object with a "text" field containing the utterance and a "speaker" field containing a unique label for the speaker.
    \item \textbf{response}: a reference response, which corresponds to the final teacher's utterance. This utterance is a JSON object with a "text" field containing the utterance and a "speaker" field containing a unique label for the speaker.
\end{itemize}

Each test sample is represented as a JSON object that uses the same format as the training sample but excludes the reference response. As a result, each test sample has two fields:

\begin{itemize}
    \item \textbf{id}: a unique identifier for the sample.
    \item \textbf{utterances}: a list of utterances, which corresponds to the preceding dialogue context. Each utterance is a JSON object with a "text" field containing the utterance and a "speaker" field containing a unique label for the speaker.
\end{itemize}

\subsection{Data Distribution and Characteristics}

The TSCC corpus is divided into three sets: train, dev, and test, each comprising 2747, 305 and 273 of the samples, respectively. 
The corpus has 3325 samples, and each sample has an average length of 7.52 turns, with about 7.33 tokens per turn on average. 
Table \ref{tab:data} presents a summary of the statistics of the TSCC corpus across the training, development, and testing sets.

\begin{table*}[ht]
\caption{Summary of the statistics for the TSCC corpus across the train, dev, and test sets.}
\label{tab:data}
\centering 
\begin{tabularx}{\textwidth}{lXXX} 
\hline
\textbf{Dataset} & \textbf{Train} & \textbf{Dev} & \textbf{Test} \\ 
\hline

Num Samples & 2747 & 305 & 273 \\ 
Avg Turns & 7.7 & 7.92 & 5.23 \\ 
Avg Tokens Per Turn & 7.29 & 7.21 & 8.27 \\ 
\hline
\end{tabularx}

\end{table*}

The TSCC corpus exhibits several characteristics that are specific to educational dialogues and pose challenges to natural language generation models. For instance, the dialogues often include technical vocabulary and idiomatic expressions related to English language learning. Additionally, the dialogues can be highly varied in terms of topic, complexity, and participant proficiency. Finally, the dialogues can include challenging responses which are based on out-of-context information, posing challenges for conversational agents. These characteristics must be taken into consideration when selecting and evaluating generative models for the TSCC corpus.

\subsection{Data Overlap and Challenges}

It is worth noting that the released development and training sets in the TSCC dataset have some overlaps, as individual conversation samples within these sets have been generated by creating chunks from larger conversations. This overlap may lead to potential biases and overfitting when training and evaluating models on this dataset. However, the test set for the BEA 2023 shared task is free of overlaps, allowing for a more accurate assessment of the model's performance in generating AI teacher responses.

The presence of overlaps in the development and training sets posed a challenge, as models inadvertently learned to predict teacher responses based on the similarities between the samples rather than genuinely understanding the context and dynamics of the teacher-student interaction. It is essential to be aware of this issue and devise strategies to mitigate the risks associated with such overlaps and ensure that the models are robust and capable of handling diverse and unseen scenarios.

To ensure the validity of our model on the validation set, we employed an iterative inclusion process to create a train-val split without any overlap between them. This process involved carefully selecting and excluding samples from the training set that had any similarity or overlap with the samples in the development set. This approach aimed to minimize the risk of data leakage and ensure that our model was evaluated on a truly unseen set of dialogues.

\section{Methods}
\label{sec:4}

The primary objective of our study is to investigate the potential of using in-context learning, supervised fine-tuning, and reinforcement learning to generate AI teacher responses in educational dialogues. Our proposed methods will be evaluated using the Teacher Student Chatroom Corpus (TSCC) dataset. In this section, we provide an overview of the three main parts of our methodology: in-context learning using GPT-4, supervised fine-tuning with existing models such as GPT-2 and DialoGPT, and supervised fine-tuning with Reinforcement Learning using the RL4LMs library \cite{ramamurthy2022reinforcement}.

\subsection{In-context Learning}

\subsubsection{GPT-4}

As a preliminary step, we investigate the potential of in-context learning using GPT-4, a state-of-the-art language model. It generates educational dialogues based on its pre-trained knowledge, which has been acquired from a vast corpus of text data during its training process (the pre-training data might have included the test set; we will address this issue in the discussion section).

To evaluate the performance of GPT-4, we prompted GPT-4 in a few-shot fashion. We retrieved 5 most similar teacher-student conversations from the TSCC dataset and provided them to the model in addition to the current conversation and instructions about the model's role as a teacher. Details about the prompt construction that helps guide the model toward generating suitable responses as a teacher can be found in the Appendix \ref{test}.

\subsection{Supervised Fine-tuning}

To further adapt pre-trained language models to the specific educational context and generate more accurate and context-aware teacher responses, we explore supervised fine-tuning using GPT-2 and DialoGPT models.

\subsubsection{GPT-2}

GPT-2 \cite{radford2019language} is a decoder-only large language model pre-trained on WebText, and we used GPT-2 Large, which has 24 transformer decoder blocks with 774 million parameters. 

We fine-tune the GPT-2 model \cite{radford2019language} using the Huggingface Library on the Teacher Student Chatroom Corpus (TSCC) dataset. For each educational dialogue, we concatenated all dialogue turns into a single string with additional information of speaker roles i.e. students or teachers. As a result, the input to the GPT-2 model consists of a sequence of text representing the conversation history, culminating in the teacher's response. We then finetuned GPT-2 Large \cite{radford2019language} with a causal language modeling task.  Details of the exact hyperparameters used during the fine-tuning process can be found in the Appendix. 

After the fine-tuning process, we evaluated the fine-tuned GPT-2 model's performance on the test set by comparing its generated teacher responses to reference responses, assessing the model's ability to generate context-aware and educationally relevant responses in line with the teacher's role in the TSCC dataset.

\subsubsection{DialoGPT}

DialoGPT \cite{zhang2019dialogpt} is a dialogue model based on the GPT-2 architecture, specifically designed for generating conversational responses. DialoGPT is trained with 147M conversation pieces extracted from Reddit \cite{zhang2019dialogpt}, and it is trained with causal language modeling objectives with multi-turn dialogue.  We adapt our training dataset with the same format as that of DialoGPT during pretraining and then prompt the DialoGPT to generate an educational dialogue of teachers in the validation set. After training, we follow the same methodology for evaluation as GPT-2 which we discussed in the earlier section.

\subsection{Supervised Fine-tuning with Reinforcement Learning}
\subsubsection{Flan-T5 Fine-tuned with RL4LMs}

To optimize the generative models for pedagogical quality, we explore the use of reinforcement learning techniques in the fine-tuning process. We employ the RL4LMs library \cite{ramamurthy2022reinforcement}, which provides an efficient and scalable framework for reinforcement learning-based language model fine-tuning.

The RL4LMs library incorporates Proximal Policy Optimization (PPO) \cite{schulman2017proximal} as the reinforcement learning algorithm, which is known for its stability and sample efficiency. The library also supports the integration of custom reward functions, allowing us to design rewards that encourage the generation of pedagogically sound teacher responses.

To implement the reinforcement learning-based fine-tuning, we first fine-tune the Flan-T5 \cite{chung2022scaling} model on the TSCC dataset using supervised learning, as described in the previous section. Next, we utilize the RL4LMs library to fine-tune the model further using the PPO algorithm. We use an equal division of the F1 as calculated by the roberta-large version of BERTScore and DialogRPT-updown as the reward function. More Details about the reinforcement learning fine-tuning process can be found in the Appendix.

The subsequent evaluation of the fine-tuned Flan-T5 model reveals the benefits of incorporating reinforcement learning into the fine-tuning process, contributing to more context-aware, relevant, and pedagogically effective AI teacher responses.

\section{Results}
\label{sec:5}

In this section, we present the results and discuss the performance of GPT-4, fine-tuned GPT-2, and fine-tuned DialoGPT models on the TSCC dataset. We analyze the strengths and weaknesses of each approach and provide insights into their potential applications and limitations in an educational context.

\begin{table*}
\caption{Validation set results} 
\centering 
\begin{tabularx}{\textwidth}{lXXX} 
\hline
 Model & BERTScore & DialogRPT
\\ [0.5ex]  
\hline   
\\ [0.5ex]  
\raisebox{1ex}{GPT-4} & \raisebox{1.5ex}{0.82} & \raisebox{1.5ex}{0.69}\\
\raisebox{1ex}{Finetuned GPT-2 Large} & \raisebox{1.5ex}{\textbf{0.94}} & \raisebox{1.5ex}{0.63}\\
\raisebox{1ex}{Finetuned DialoGPT Large} & \raisebox{1.5ex}{\textbf{0.94}} & \raisebox{1.5ex}{0.64} \\
\raisebox{1ex}{Finetuned Flan-T5 w/ RL} & \raisebox{1.5ex}{0.89} & \raisebox{1.5ex}{\textbf{0.71}}
\\
\hline 
\end{tabularx}  
\end{table*}

\begin{table*}
\caption{Test set results} 
\centering 
\begin{tabularx}{\textwidth}{lXXX} 
\hline
 Model & BERTScore & DialogRPT
\\ [0.5ex]  
\hline   
\\ [0.5ex]  
\raisebox{1ex}{GPT-4} & \raisebox{1.5ex}{\textbf{0.8}} & \raisebox{1.5ex}{\textbf{0.70}}\\
\raisebox{1ex}{Finetuned Flan-T5 w/ RL} & \raisebox{1.5ex}{0.66} & \raisebox{1.5ex}{0.34}
\\
\hline 
\end{tabularx}  
\end{table*}  

\subsection{GPT-4}

The GPT-4 model, without fine-tuning on the TSCC dataset, demonstrates a relatively strong performance in generating educational dialogues. The generated teacher responses are generally fluent and contextually relevant, indicating that GPT-4 has a good understanding of the educational context based on its pre-trained knowledge. However, some limitations are observed in the model's ability to generate accurate and pedagogically sound responses consistently.

The carefully crafted prompt provided to the model plays a crucial role in guiding GPT-4 toward generating suitable responses as a teacher. Although the model is capable of generating contextually relevant and linguistically correct responses, it may not always produce the most pedagogically sound or helpful responses for the students. This limitation highlights the importance of fine-tuning the model on a specific educational dataset, such as TSCC, to further enhance its performance in generating AI teacher responses.

Additionally, due to the nature of the dataset, where conversations were often cut off, the model sometimes lacked the full context needed to generate meaningful responses that accurately represented the ground truth. Despite this limitation, GPT-4's responses were generally sensible and appropriate given the available context.

\subsection{Finetuned GPT-2}
We observe that compared with DialoGPT, GPT-2 usually generates longer and more formal responses, even with the same generation hyperparameters. 

\subsection{Finetuned DialoGPT}
We observe that DialoGPT usually generates shorter and more vernacular responses. It fits better in a conversational setting, but sometimes the educational uptakes are not satisfactory since the responses are not guiding students to learn the language.

\subsection{Finetuned Flan-T5 w/ RL}
We observe that the results of Flan-T5 w/ RL on the validation set are really good suggesting that the model was able to hack the metrics designed as the reward. On the contrary, it is performing poorly on the test set suggesting that it overfits the validation set. We hypothesize two reasons for this to be the case:
the way conversations are split into chunks in the training dataset or the difference in distribution between the training set and the test set.

\section{Discussion}
Conversational agents have the potential to revolutionize the teaching landscape by addressing several challenges and enhancing the overall learning experience for both students and educators \cite{wollny2021we}. However, developing conversational agents that can behave like human teachers requires addressing several challenges \cite{tack2022ai}. 
\newline

\noindent\textbf{Data challenges.} As noted in the subsections above, the generations from the GPT-$4$ model outperformed all the fine-tuned models, with and without reinforcement learning. To this end,
we put forward the proposition that an array of dataset features plays a crucial role in posing significant challenges to the fine-tuning process of generative models. These features include several dataset characteristics, including sampling, representativeness, prompt and response lengths, and dialogue completeness---upon manual inspection, we identified several dialogues to be cut off---pose serious challenges in achieving superior performance with fine-tuning. Furthermore, upon random inspection of the generations from the fine-tuned models, we identified that these models seem to have learned simple, generic, often inappropriate yet correct responses such as ``thank you'' and ``okay.'' While more recent language models have been shown to have high few-shot performance, we believe that fine-tuned models could adapt better to provide domain-specific responses in comparison. To achieve this, we emphasize the need for extending the current dataset to include longer prompts with more context.
\newline
\begin{figure}[ht]
    \begin{tabular}{ll}
        [student] & someone plugged the charger in \\\relax
        [teacher] & that's bad, charger must be $\_\_\_$? \\\relax
        [student] & umm $\dots$ \\\relax 
        \\[-10pt]\hline\\[-10pt]
        
        [model] & (a) \textcolor{red}{plugged in} $\leftarrow$ score: $0.91$ \tabularnewline
        & (b) \textcolor{ForestGreen}{disconnected} $\leftarrow$ score: $0.90$ \tabularnewline
        \\[-10pt]\hline\\[-10pt]

        [reference] & plugged out \tabularnewline
    \end{tabular}
    \caption{An example dialog demonstrating that two opposing responses, (a) and (b), ranked alike using the BERTScore metric.}
    \label{fig:discussion}
\end{figure}
It is important to acknowledge that these models might not be as effective as desired in their response generation due to these intricacies. The current efforts made by the research community to collect and build quality datasets encompassing enough information about the educational task to enable AI teacher generative models to fully generalize in any context is what we assess to be the main focus that the community should adopt \cite{edpost}.
\newline
\newline
\noindent\textbf{Evaluation metrics.} 
In addition, we emphasize that to truly gauge the efficiency of these AI-powered teaching models, it is vital to go a step further and examine their ability to comprehend the unique nuances in the students' queries and cater to their particular educational requirements. This implies the need for a pedagogically meaningful evaluation metric. We believe that it is crucial for the research community to embrace this as the second primary focus.
While common evaluation metrics such as BERTScore and DialogRPT are commonly used in several language and dialog modeling tasks, it is important to note that these metrics were not fundamentally designed to capture the level of pedagogical meaningfulness in the generated responses. As an example, consider the dialog shown in Figure~\ref{fig:discussion}---depending on the given context, only one of the responses (option (b): disconnected) is correct, while both the responses are ranked as equally correct by the BERTScore metric. Commonly-used domain-agnostic metrics often serve as a proxy for how coherent and human-like the generated responses are. However, for more goal-oriented tasks such as modeling teacher-student conversational dialogues, these metrics seem to fall short. This generalization gap becomes more apparent on analyzing the results from the fine-tuned Flan-T$5$ model with a feedback loop based on BERTScore and DialogRPT scores---despite the model performing significantly well on training and validation sets, it failed to generalize on unseen test data.
In an effort to advance research on this front, we note the need for auxiliary training-level metrics, including the faithfulness of the generation to the true response, to ensure that the generations are context-aware and factually accurate (e.g., correct option (b) vs. incorrect option (a) in Figure~\ref{fig:discussion}).
\newline
\newline
\noindent\textbf{GPT-4 unknown pre-training data.} We understand that the use of GPT-4 as a baseline in our study presents challenges due to its unknown training data. Yet, whether GPT-4 has seen parts of the TSCC dataset during its pre-training or not, the improvement of performance compared to the reference with regard to the DialogRPT scores and human evaluation scores attached to the leaderboard of the shared task suggests that the potential of using such high-performing models in this domain warrants further exploration.

\section{Conclusion}
\label{sec:6}

In this paper, we explored the potential of using large pre-trained language models and reinforcement learning for generating AI teacher responses in an educational context. We first presented a few-shot approach using the GPT-4 model, which demonstrated promising results in generating contextually relevant and fluent responses, but with limitations in generating pedagogically sound responses consistently. We then fine-tuned GPT-2 and DialoGPT on the TSCC dataset and evaluated their performance using BERTScore and DialogRPT metrics. We also proposed an approach using RL to optimize directly for pedagogical values. We hypothesized that several dataset characteristics (e.g., dialog completeness, sampling) pose challenges to achieving superior performance with fine-tuning. To this end, we recommend the extension of the dataset to include longer prompts with extended context. Finally, we also draw attention to the need for more domain-specific metrics (in both evaluation and reward-based training) in enabling the generation of accurate, context-aware, and factually correct teacher responses.

\bibliography{custom}

\begin{thebibliography}{33}
\expandafter\ifx\csname natexlab\endcsname\relax\def\natexlab#1{#1}\fi

\bibitem[{Bibauw et~al.(2022)Bibauw, Fran{\c{c}}ois, and
  Desmet}]{bibauw2022dialogue}
Serge Bibauw, Thomas Fran{\c{c}}ois, and Piet Desmet. 2022.
\newblock Dialogue systems for language learning: Chatbots and beyond.
\newblock In \emph{The Routledge handbook of second language acquisition and
  technology}, pages 121--135. Routledge.

\bibitem[{Bommasani et~al.(2021)Bommasani, Hudson, Adeli, Altman, Arora, von
  Arx, Bernstein, Bohg, Bosselut, Brunskill
  et~al.}]{bommasani2021opportunities}
Rishi Bommasani, Drew~A Hudson, Ehsan Adeli, Russ Altman, Simran Arora, Sydney
  von Arx, Michael~S Bernstein, Jeannette Bohg, Antoine Bosselut, Emma
  Brunskill, et~al. 2021.
\newblock On the opportunities and risks of foundation models.
\newblock \emph{arXiv preprint arXiv:2108.07258}.

\bibitem[{Brown et~al.(2020)Brown, Mann, Ryder, Subbiah, Kaplan, Dhariwal,
  Neelakantan, Shyam, Sastry, Askell et~al.}]{brown2020language}
Tom Brown, Benjamin Mann, Nick Ryder, Melanie Subbiah, Jared~D Kaplan, Prafulla
  Dhariwal, Arvind Neelakantan, Pranav Shyam, Girish Sastry, Amanda Askell,
  et~al. 2020.
\newblock Language models are few-shot learners.
\newblock \emph{Advances in neural information processing systems},
  33:1877--1901.

\bibitem[{Caines et~al.(2022)Caines, Yannakoudakis, Allen, P{\'e}rez-Paredes,
  Byrne, and Buttery}]{caines2022teacher}
Andrew Caines, Helen Yannakoudakis, Helen Allen, Pascual P{\'e}rez-Paredes,
  Bill Byrne, and Paula Buttery. 2022.
\newblock The teacher-student chatroom corpus version 2: more lessons, new
  annotation, automatic detection of sequence shifts.
\newblock In \emph{Proceedings of the 11th Workshop on NLP for Computer
  Assisted Language Learning}, pages 23--35.

\bibitem[{Caines et~al.(2020)Caines, Yannakoudakis, Edmondson, Allen,
  P{\'e}rez-Paredes, Byrne, and Buttery}]{caines2020teacher}
Andrew Caines, Helen Yannakoudakis, Helena Edmondson, Helen Allen, Pascual
  P{\'e}rez-Paredes, Bill Byrne, and Paula Buttery. 2020.
\newblock The teacher-student chatroom corpus.
\newblock \emph{arXiv preprint arXiv:2011.07109}.

\bibitem[{Chang et~al.(2020)Chang, Chiam, Fu, Wang, Zhang, and
  Danescu-Niculescu-Mizil}]{chang2020convokit}
Jonathan~P Chang, Caleb Chiam, Liye Fu, Andrew~Z Wang, Justine Zhang, and
  Cristian Danescu-Niculescu-Mizil. 2020.
\newblock Convokit: A toolkit for the analysis of conversations.
\newblock \emph{arXiv preprint arXiv:2005.04246}.

\bibitem[{Chen et~al.(2019)Chen, Ferreira, Lang, and
  Gasevic}]{chen2019predictors}
Guanliang Chen, Rafael Ferreira, David Lang, and Dragan Gasevic. 2019.
\newblock Predictors of student satisfaction: A large-scale study of
  human-human online tutorial dialogues.
\newblock \emph{International Educational Data Mining Society}.

\bibitem[{Chung et~al.(2022)Chung, Hou, Longpre, Zoph, Tay, Fedus, Li, Wang,
  Dehghani, Brahma et~al.}]{chung2022scaling}
Hyung~Won Chung, Le~Hou, Shayne Longpre, Barret Zoph, Yi~Tay, William Fedus,
  Eric Li, Xuezhi Wang, Mostafa Dehghani, Siddhartha Brahma, et~al. 2022.
\newblock Scaling instruction-finetuned language models.
\newblock \emph{arXiv preprint arXiv:2210.11416}.

\bibitem[{Cohen et~al.(2020)Cohen, Wong, Krishnamachari, and
  Berlin}]{cohen2020teacher}
Julie Cohen, Vivian Wong, Anandita Krishnamachari, and Rebekah Berlin. 2020.
\newblock Teacher coaching in a simulated environment.
\newblock \emph{Educational evaluation and policy analysis}, 42(2):208--231.

\bibitem[{Collins(1982)}]{collins1982discourse}
James Collins. 1982.
\newblock Discourse style, classroom interaction and differential treatment.
\newblock \emph{Journal of reading behavior}, 14(4):429--437.

\bibitem[{Demszky et~al.(2021)Demszky, Liu, Mancenido, Cohen, Hill, Jurafsky,
  and Hashimoto}]{demszky2021measuring}
Dorottya Demszky, Jing Liu, Zid Mancenido, Julie Cohen, Heather Hill, Dan
  Jurafsky, and Tatsunori Hashimoto. 2021.
\newblock Measuring conversational uptake: A case study on student-teacher
  interactions.
\newblock \emph{arXiv preprint arXiv:2106.03873}.

\bibitem[{Jaques et~al.(2019)Jaques, Ghandeharioun, Shen, Ferguson, Lapedriza,
  Jones, Gu, and Picard}]{jaques2019way}
Natasha Jaques, Asma Ghandeharioun, Judy~Hanwen Shen, Craig Ferguson, Agata
  Lapedriza, Noah Jones, Shixiang Gu, and Rosalind Picard. 2019.
\newblock Way off-policy batch deep reinforcement learning of implicit human
  preferences in dialog.
\newblock \emph{arXiv preprint arXiv:1907.00456}.

\bibitem[{Jarratt(2023)}]{edpost}
Daniel Jarratt. 2023.
\newblock \href
  {https://www.edpost.com/stories/education-data-what-gpt-needs-to-work-for-kids}
  {Chatgpt: The double-edged sword of ai in education}.

\bibitem[{Kane et~al.(2022)Kane, Hill, and Staiger}]{Kane2022}
Thomas Kane, Heather Hill, and Douglas Staiger. 2022.
\newblock \href {https://doi.org/10.3886/ICPSR36095.v4} {National center for
  teacher effectiveness main study}.

\bibitem[{Kasneci et~al.(2023)Kasneci, Se{\ss}ler, K{\"u}chemann, Bannert,
  Dementieva, Fischer, Gasser, Groh, G{\"u}nnemann, H{\"u}llermeier
  et~al.}]{kasneci2023chatgpt}
Enkelejda Kasneci, Kathrin Se{\ss}ler, Stefan K{\"u}chemann, Maria Bannert,
  Daryna Dementieva, Frank Fischer, Urs Gasser, Georg Groh, Stephan
  G{\"u}nnemann, Eyke H{\"u}llermeier, et~al. 2023.
\newblock Chatgpt for good? on opportunities and challenges of large language
  models for education.
\newblock \emph{Learning and Individual Differences}, 103:102274.

\bibitem[{Norouzi et~al.(2016)Norouzi, Bengio, Jaitly, Schuster, Wu, Schuurmans
  et~al.}]{norouzi2016reward}
Mohammad Norouzi, Samy Bengio, Navdeep Jaitly, Mike Schuster, Yonghui Wu, Dale
  Schuurmans, et~al. 2016.
\newblock Reward augmented maximum likelihood for neural structured prediction.
\newblock \emph{Advances In Neural Information Processing Systems}, 29.

\bibitem[{Pasunuru and Bansal(2018)}]{pasunuru2018multi}
Ramakanth Pasunuru and Mohit Bansal. 2018.
\newblock Multi-reward reinforced summarization with saliency and entailment.
\newblock \emph{arXiv preprint arXiv:1804.06451}.

\bibitem[{Radford et~al.(2019)Radford, Wu, Child, Luan, Amodei, Sutskever
  et~al.}]{radford2019language}
Alec Radford, Jeffrey Wu, Rewon Child, David Luan, Dario Amodei, Ilya
  Sutskever, et~al. 2019.
\newblock Language models are unsupervised multitask learners.
\newblock \emph{OpenAI blog}, 1(8):9.

\bibitem[{Raffel et~al.(2020)Raffel, Shazeer, Roberts, Lee, Narang, Matena,
  Zhou, Li, and Liu}]{raffel2020exploring}
Colin Raffel, Noam Shazeer, Adam Roberts, Katherine Lee, Sharan Narang, Michael
  Matena, Yanqi Zhou, Wei Li, and Peter~J Liu. 2020.
\newblock Exploring the limits of transfer learning with a unified text-to-text
  transformer.
\newblock \emph{The Journal of Machine Learning Research}, 21(1):5485--5551.

\bibitem[{Ramamurthy et~al.(2022)Ramamurthy, Ammanabrolu, Brantley, Hessel,
  Sifa, Bauckhage, Hajishirzi, and Choi}]{ramamurthy2022reinforcement}
Rajkumar Ramamurthy, Prithviraj Ammanabrolu, Kiant{\'e} Brantley, Jack Hessel,
  Rafet Sifa, Christian Bauckhage, Hannaneh Hajishirzi, and Yejin Choi. 2022.
\newblock Is reinforcement learning (not) for natural language processing?:
  Benchmarks, baselines, and building blocks for natural language policy
  optimization.
\newblock \emph{arXiv preprint arXiv:2210.01241}.

\bibitem[{Ranzato et~al.(2015)Ranzato, Chopra, Auli, and
  Zaremba}]{ranzato2015sequence}
Marc'Aurelio Ranzato, Sumit Chopra, Michael Auli, and Wojciech Zaremba. 2015.
\newblock Sequence level training with recurrent neural networks.
\newblock \emph{arXiv preprint arXiv:1511.06732}.

\bibitem[{Roller et~al.(2020)Roller, Dinan, Goyal, Ju, Williamson, Liu, Xu,
  Ott, Shuster, Smith et~al.}]{roller2020recipes}
Stephen Roller, Emily Dinan, Naman Goyal, Da~Ju, Mary Williamson, Yinhan Liu,
  Jing Xu, Myle Ott, Kurt Shuster, Eric~M Smith, et~al. 2020.
\newblock Recipes for building an open-domain chatbot.
\newblock \emph{arXiv preprint arXiv:2004.13637}.

\bibitem[{Scao et~al.(2022)Scao, Fan, Akiki, Pavlick, Ili{\'c}, Hesslow,
  Castagn{\'e}, Luccioni, Yvon, Gall{\'e} et~al.}]{scao2022bloom}
Teven~Le Scao, Angela Fan, Christopher Akiki, Ellie Pavlick, Suzana Ili{\'c},
  Daniel Hesslow, Roman Castagn{\'e}, Alexandra~Sasha Luccioni, Fran{\c{c}}ois
  Yvon, Matthias Gall{\'e}, et~al. 2022.
\newblock Bloom: A 176b-parameter open-access multilingual language model.
\newblock \emph{arXiv preprint arXiv:2211.05100}.

\bibitem[{Schulman et~al.(2017)Schulman, Wolski, Dhariwal, Radford, and
  Klimov}]{schulman2017proximal}
John Schulman, Filip Wolski, Prafulla Dhariwal, Alec Radford, and Oleg Klimov.
  2017.
\newblock Proximal policy optimization algorithms.
\newblock \emph{arXiv preprint arXiv:1707.06347}.

\bibitem[{Serban et~al.(2017)Serban, Sankar, Germain, Zhang, Lin, Subramanian,
  Kim, Pieper, Chandar, Ke et~al.}]{serban2017deep}
Iulian~V Serban, Chinnadhurai Sankar, Mathieu Germain, Saizheng Zhang, Zhouhan
  Lin, Sandeep Subramanian, Taesup Kim, Michael Pieper, Sarath Chandar,
  Nan~Rosemary Ke, et~al. 2017.
\newblock A deep reinforcement learning chatbot.
\newblock \emph{arXiv preprint arXiv:1709.02349}.

\bibitem[{Shen et~al.(2015)Shen, Cheng, He, He, Wu, Sun, and
  Liu}]{shen2015minimum}
Shiqi Shen, Yong Cheng, Zhongjun He, Wei He, Hua Wu, Maosong Sun, and Yang Liu.
  2015.
\newblock Minimum risk training for neural machine translation.
\newblock \emph{arXiv preprint arXiv:1512.02433}.

\bibitem[{Tack et~al.(2023)Tack, Kochmar, Yuan, Bibauw, and
  Piech}]{tack_bea_2023}
Ana{\"i}s Tack, Ekaterina Kochmar, Zheng Yuan, Serge Bibauw, and Chris Piech.
  2023.
\newblock The {{BEA}} 2023 {{Shared Task}} on {{Generating AI Teacher
  Responses}} in {{Educational Dialogues}}.
\newblock In \emph{Proceedings of the 18th {{Workshop}} on {{Innovative Use}}
  of {{NLP}} for {{Building Educational Applications}}}, page to appear,
  {Toronto, Canada}. {Association for Computational Linguistics}.

\bibitem[{Tack and Piech(2022)}]{tack2022ai}
Ana{\"\i}s Tack and Chris Piech. 2022.
\newblock The ai teacher test: Measuring the pedagogical ability of blender and
  gpt-3 in educational dialogues.
\newblock \emph{arXiv preprint arXiv:2205.07540}.

\bibitem[{Team(2022)}]{team2022chatgpt}
OpenAI Team. 2022.
\newblock Chatgpt: Optimizing language models for dialogue.

\bibitem[{Touvron et~al.(2023)Touvron, Lavril, Izacard, Martinet, Lachaux,
  Lacroix, Rozi{\`e}re, Goyal, Hambro, Azhar et~al.}]{touvron2023llama}
Hugo Touvron, Thibaut Lavril, Gautier Izacard, Xavier Martinet, Marie-Anne
  Lachaux, Timoth{\'e}e Lacroix, Baptiste Rozi{\`e}re, Naman Goyal, Eric
  Hambro, Faisal Azhar, et~al. 2023.
\newblock Llama: Open and efficient foundation language models.
\newblock \emph{arXiv preprint arXiv:2302.13971}.

\bibitem[{Wollny et~al.(2021)Wollny, Schneider, Di~Mitri, Weidlich, Rittberger,
  and Drachsler}]{wollny2021we}
Sebastian Wollny, Jan Schneider, Daniele Di~Mitri, Joshua Weidlich, Marc
  Rittberger, and Hendrik Drachsler. 2021.
\newblock Are we there yet?-a systematic literature review on chatbots in
  education.
\newblock \emph{Frontiers in artificial intelligence}, 4:654924.

\bibitem[{Zhang et~al.(2022)Zhang, Roller, Goyal, Artetxe, Chen, Chen, Dewan,
  Diab, Li, Lin et~al.}]{zhang2022opt}
Susan Zhang, Stephen Roller, Naman Goyal, Mikel Artetxe, Moya Chen, Shuohui
  Chen, Christopher Dewan, Mona Diab, Xian Li, Xi~Victoria Lin, et~al. 2022.
\newblock Opt: Open pre-trained transformer language models.
\newblock \emph{arXiv preprint arXiv:2205.01068}.

\bibitem[{Zhang et~al.(2019)Zhang, Sun, Galley, Chen, Brockett, Gao, Gao, Liu,
  and Dolan}]{zhang2019dialogpt}
Yizhe Zhang, Siqi Sun, Michel Galley, Yen-Chun Chen, Chris Brockett, Xiang Gao,
  Jianfeng Gao, Jingjing Liu, and Bill Dolan. 2019.
\newblock Dialogpt: Large-scale generative pre-training for conversational
  response generation.
\newblock \emph{arXiv preprint arXiv:1911.00536}.

\end{thebibliography}
\bibliographystyle{acl_natbib}

\clearpage

\appendix

\section{Appendix}
\label{test}

\subsection{GPT-4 Prompt Construction}

To evaluate the performance of GPT-4, we provided it with a few-shot prompt that includes a selection of similar teacher-student conversations from the TSCC dataset. This approach helps guide the model toward generating suitable responses as a teacher. The prompt is constructed as follows:

\begin{itemize}
    \item We direct the system role to act as a teacher and encourage learning by using the prompt as given below.
    \item Retrieve the 5 most similar teacher-student conversations from the TSCC dataset. This is done by computing the cosine similarity between the input conversation context and the current conversation context in the dataset using embeddings generated by the text-embedding-ada-002 model.
    \item Concatenate the selected conversations with the input conversation, separated by special tokens to indicate the beginning and end of a new sample conversation.
\end{itemize}

This prompt construction aims to provide GPT-4 with the necessary context and guidance to generate accurate and pedagogically relevant responses in the context of teacher-student dialogues. The prompt is designed as follows:

\begin{quote}
    You are acting as a teacher, and you are helping a student learn. Be patient, helpful, and kind. Don't be superimposing; give short responses to encourage learning. Make the student feel comfortable and confident, and help them learn. Now, join the following conversation: <conversation context>
\end{quote}

The prompt is designed using the following directives in mind:-
\begin{itemize}
    \item We instruct the system with several indicators to act as a teacher and provide helpful advice to the student. 
    \item To mitigate the challenge of generating teacher-like responses, we advise the model to be patient, kind, and helpful to the student. 
    \item Through the directive to keep responses short and encouraging, we guide the model toward generating suitable responses that might help the student learn effectively. 
    \item The model is also instructed to make the student feel comfortable and confident in their learning process, providing an overall supportive environment for the student. 
    \item Finally, the conversation context is provided to the model to set the context for the given student query, allowing the model to generate appropriate responses given the conversation context. 
\end{itemize}

Combining all these aspects together, we aim to guide the model toward generating contextually relevant and pedagogically meaningful responses in the given teacher-student dialogue.

We use the following hyperparameters for querying the GPT-4 model:
\begin{itemize}
    \item Model: gpt-4-0314
    \item Temperature: 1
    \item Max Tokens: 100
    \item Top p: 1
\end{itemize}

\subsection{Fine-tuning Exact Parameters}

For our supervised fine-tuning experiments, we used the following hyperparameters:

\subsubsection{GPT-2}
\begin{itemize}
    \item Learning rate: 1e-5
    \item Batch size: 32
    \item Epochs: 10
    \item Max sequence length: 1024
    \item Optimizer: AdamW
    \item Scheduler: linear learning rate scheduler
\end{itemize}

\subsubsection{DialoGPT}
\begin{itemize}
    \item Learning rate: 1e-5
    \item Batch size: 32
    \item Epochs: 10
    \item Max sequence length: 1024
    \item Optimizer: AdamW
    \item Scheduler: linear learning rate scheduler
\end{itemize}

\subsection{Supervised Fine-tuning with Reinforcement Learning Details}

To implement the reinforcement learning-based fine-tuning using the RL4LMs library, we first fine-tuned the Flan-T5 model on the TSCC dataset using supervised learning. After this initial fine-tuning step, we utilized the RL4LMs library to fine-tune the model further using reinforcement learning. We used an equal division of the BERTScore and DialogRPT as the reward function to optimize the model for pedagogical quality. The following hyperparameters were used for the reinforcement learning fine-tuning process:

\begin{itemize}
    \item Learning rate: 1e-6
    \item Batch size: 64
    \item Epochs: 5
    \item Max prompt length: 512
    \item Max episode length: 100
    \item Optimizer: AdamW
    \item Scheduler: linear learning rate scheduler
\end{itemize}

The YAML file for the RL4LMs scipt is as follows:
\begin{minted}[
    gobble=4
  ]{yaml}
    tokenizer:
  model_name: google/flan-t5-small
  padding_side: left
  truncation_side: left
  pad_token_as_eos_token: False
reward_fn:
  id: dialog_rpt_bert
  args:
    BERTScore_coeff: 0.5
    DialogRPT_coeff: 0.5
datapool:
  id: bea
  truncate: False
  args: {}
env:
  n_envs: 1
  args:
    max_prompt_length: 100
    max_episode_length: 20
    terminate_on_eos: True
    context_start_token: 0
    prompt_truncation_side: "right"
alg:
  id: ppo_separate
  args:
    n_steps: 20
    batch_size: 64
    verbose: 1
    learning_rate: 0.000001
    clip_range: 0.2
    n_epochs: 1
    value_update_epochs: 3
    # batchify: False
    gae_lambda: 0.95
    gamma: 0.99
    ent_coef: 0.01
  kl_div:
    coeff: 0.001
    target_kl: 2.0
  policy:
    id: seq2seq_lm_actor_critic_policy
    args:
      model_name: google/flan-t5-small
      apply_model_parallel: True
      prompt_truncation_side: "right"
      generation_kwargs:
        do_sample: True
        top_k: 0
        min_length: 9
        max_new_tokens: 20
train_evaluation:
  eval_batch_size: 64
  n_iters: 200
  eval_every: 20
  save_every: 10
  metrics:
    - id: bert_score
      args:
        language: en
    - id: dialog_rpt
      args:
        model_name: "microsoft/DialogRPT
                    -updown"
        label_ix: 0
        batch_size: 1
    # - id: uptake
    #   args:
    #     model_name: None
    #     label_ix: 0
    #     batch_size: 1
  generation_kwargs:
    num_beams: 5
    min_length: 9
    max_new_tokens: 20
  \end{minted}

\end{document}